\documentclass[11pt]{article}

\usepackage[final]{acl}

\usepackage{times}
\usepackage{latexsym}
\usepackage[T1]{fontenc}
\usepackage[utf8]{inputenc}
\usepackage{microtype}
\usepackage{inconsolata}
\usepackage{graphicx}
\usepackage{booktabs}
\usepackage{multirow}
\usepackage[most]{tcolorbox}
\usepackage{amssymb}

\newcommand{\sysname}{Hybrid memory}    
\newcommand{\epimem}{episodic memory}        
\newcommand{\structmem}{structured relational memory}

\newcommand{\locomo}{LoCoMo}
\newcommand{\longmemeval}{LongMemEval-S}

\title{Long-Term Agent Memory through Complementary Curation and Consolidation}

\author{
  \textbf{Abhishek Gupta \quad Dongge Han \quad Daniel Madrigal Diaz\quad Wei-Ning Chen} \\
  \textbf{Mayukh Das \quad Victor Rühle \quad Saravan Rajmohan} \\[2pt]
  Microsoft \\
}

\begin{document}
\maketitle

\begin{abstract}
Long-horizon dialogue agents require effective memory systems to recall facts, events, and relationships
from interaction histories. In this work, we propose a hybrid-memory architecture comprising \emph{episodic memories} and \emph{structured relational memories} and a consolidation mechanism which improves efficiency. 
We conduct an empirical study across two benchmarks:
\locomo{} and \longmemeval{}.
Our experiments demonstrate that the two memory types are consistently
complementary, combining both outperforms state-of-the-art memory frameworks and achieves 90.2\% accuracy on \locomo{} and 85.5\% on \longmemeval{}.
\end{abstract}

\section{Introduction}
Recent advances in large language models (LLMs) have enabled agentic capability of sustaining multi-turn interactions~\cite{wang2024survey, hu-etal-2025-os}. However, their ability to recall, integrate, and reason over long interaction histories remains fundamentally limited. 
These limitations have motivated the development of long-term memory systems that explicitly store and retrieve learning from past interactions~\cite{mem_intro}. By enabling selective access to historical experience, these mechanisms mitigate the need for massive context windows, fostering more coherent and efficient agent behavior.

Prior work on agentic memories can be primarily categorized into two types. \emph{Parametric memory}~\cite{mlpmem, selfupdatable, elder} embeds knowledge directly into model parameters through training, enabling low-latency inference; however, encoding is computationally expensive, frequent updates are infeasible, and unconstrained modifications may induce hallucination. In contrast, \emph{non-parametric memory}~\cite{chhikara2025mem0, rasmussen2025zep, nan2025nemori} leverages external memory systems that store and retrieve information on demand, offering scalability and adaptability without training, though at the expense of additional retrieval latency.

In this work, we focus on the latter approach, and propose a hybrid memory system of \textbf{episodic memory} which stores distinct events from the interactions, preserving the temporal ordering and contextual richness, and
\textbf{structured relational memory} which extracts entity specific tuples from dialogues, making implicit connections between people, places, and facts explicit and enabling relational inference across sessions.
These two representations reflect fundamentally different assumptions about
what makes memory useful: episodic memory preserves \emph{narrative context
and sequence}, while structured relational memory prioritizes \emph{compact,
queryable relational structure}. 
Furthermore, we introduce an offline \textbf{memory consolidation mechanism} that selectively merges related or redundant memories and removes conflicting or outdated entries, enabling scalable and efficient long-term memory management.

We evaluate our system on two long-horizon dialogue benchmarks,
\locomo{}~\cite{maharana2024evaluating} and
\longmemeval{}~\cite{wu2024longmemeval}.

Our empirical results demonstrate that episodic and structured relational memory together deliver strong empirical performance, with our hybrid system outperforming state-of-the-art baseline methods on both benchmarks. Our memory consolidation method further shows effective compression of the memory store by 16.34\% without much loss in performance. Moreover, on the \locomo{} benchmark, the consolidated system improves accuracy on both single-hop and multi-hop QA tasks.

\begin{figure*}[t] 
    \centering
    \includegraphics[width=\textwidth]{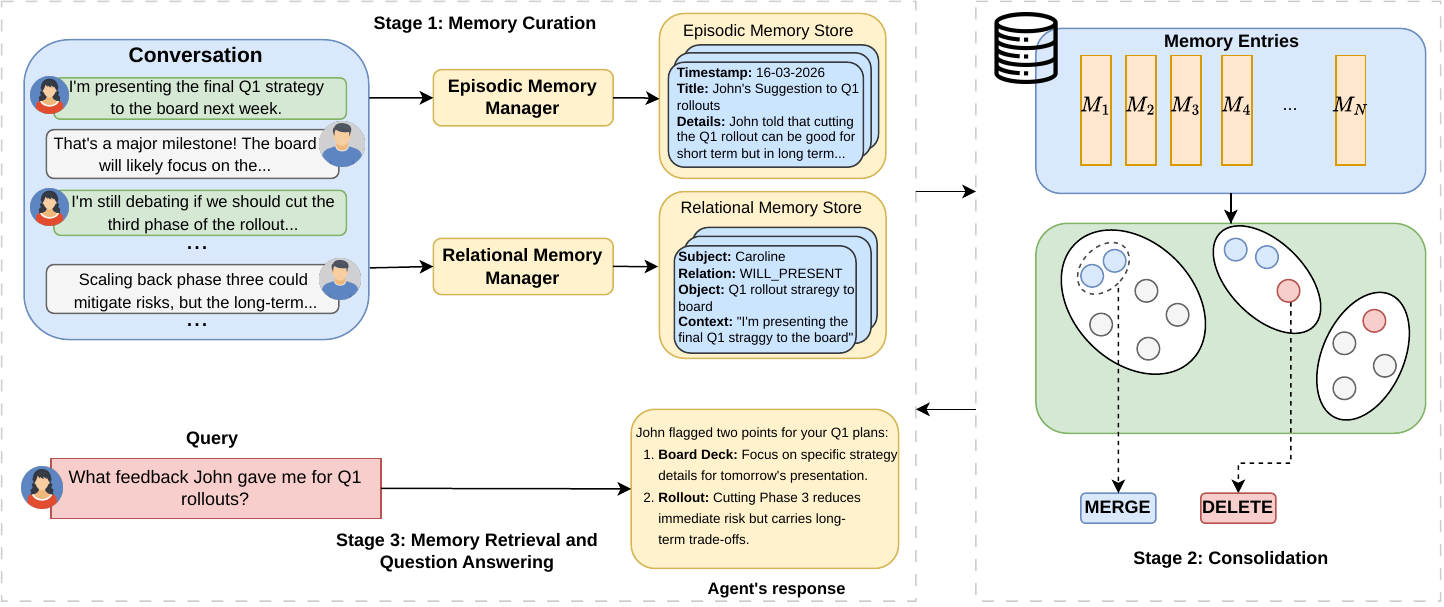}
    \caption{Architecture Overview. \textbf{Stage 1 (Memory Curation):} The Episodic Memory Manager stores timestamped summaries, while the Relational Memory Manager extracts structured knowledge triplets with conversational context. \textbf{Stage 2 (Consolidation):} Memory entries are clustered based on semantic similarity, with memories merged or deleted to maintain efficiency. \textbf{Stage 3 (Memory Retrieval and QA):} Given a user query, relevant memories are retrieved from both episodic and relational stores to generate contextually grounded responses.}
    \vspace{-5mm}
    \label{fig:architecture}
\end{figure*}

\section{Related Work}
\label{sec:related}
Recent work on long-term memory includes \emph{virtual context management}: \textsc{MemoryOS}~\cite{memos}, \textsc{MemGPT}~\cite{packer2023memgpt}, and \textsc{ReadAgent}~\cite{readagent} page information between a context window and external storage to handle inputs that exceed the model's attention span. A second line of work extracts and maintains salient content across sessions: \textsc{Mem0}~\cite{chhikara2025mem0} applies per-turn CRUD operations over extracted facts, \textsc{A-Mem}~\cite{amem} organizes memories as evolving Zettelkasten notes with adaptive linking, and \textsc{H-Mem}~\cite{hmem} adds hierarchical indexing for scalable routing. \textsc{Zep}~\cite{rasmussen2025zep} builds a temporal graph to support structured multi-hop traversal, while \textsc{Nemori}~\cite{nan2025nemori} forms coherent episodes through adaptive distillation.

Our work differs in two ways. First, we combine \emph{episodic memory} with a lightweight \emph{relational tuple store} retrieved by semantic similarity and grounded with timestamps and source context, leveraging complementary narrative and entity-relation memories. Second, our \emph{offline consolidation} phase merges redundant memories, improving efficiency without much loss in accuracy. A detailed comparison is provided in Appendix~\ref{app:extended_related}.

\section{Method}
\label{sec:method}

\sysname{} operates in three phases: an offline \emph{curation} phase that
processes each conversation session, an offline consolidation phase, and a final \emph{retrieval-and-generation} phase that responds to user queries.
Figure~\ref{fig:architecture} presents an overview of our proposed memory system.

\subsection{Memory Curation}
\label{sec:curation}

Both memory stores are populated at session granularity: after each
conversation session concludes, we run curation over its transcript.

\paragraph{Episodic Memory.}

For each session, we create holistic episodic memories using an LLM. As shown in \autoref{fig:architecture}, each entry contains three fields:
a short \textbf{title} identifying the episode, a one-sentence \textbf{summary} distilling its key point, and a \textbf{details} paragraph preserving the narrative context, including the event and any relevant background. Each episodic record is additionally annotated with the session timestamp. The prompt for episodic memory curation is shown in Appendix~\ref{app:episodic_memory_prompt}.

\paragraph{Structured Relational Memory.}
We observe that conversations naturally express persistent facts about entities and their relationships, including who someone is, what they do, who they are connected to. These relationships are better represented as structured relational tuples. To this end, we prompt an LLM to extract entity-relation tuples from the dialogues.
Each tuple is stored as $(e,\, r,\, o,\, t,\, c)$. As shown in \autoref{fig:architecture},
$e$ is the \textbf{\emph{entity}}, $r$ the \textbf{\emph{relation label}}, $o$ the \textbf{\emph{object}},
$t$ the \textbf{\emph{timestamp}}, and $c$ the \textbf{\emph{source context}} which captures the richer context from which the tuple was extracted. The contextual description $c$ provides provenance: it records not just
\emph{what} the tuple asserts but \emph{in what context} it was mentioned,
which is important for preserving nuance that structured extraction
inevitably loses, such as the speaker's intent, the circumstances under which a fact was stated, and additional details. The timestamp $t$ additionally grounds each triplet temporally, addressing a common failure mode of relation-based memory where relational facts are stored without any record of when they were true.
The prompt for relational memory curation is shown in Appendix~\ref{app:relational_memory_prompt}.

\begin{table*}[t]
\centering
\small
\vspace{2mm}
\resizebox{\textwidth}{!}{
\begin{tabular}{l|l|ccc|ccc|ccc|ccc|ccc}
\toprule
\multirow{2}{*}{\textbf{Model}} & \multirow{2}{*}{\textbf{Method}} & \multicolumn{3}{c|}{\textbf{Single-Hop}} & \multicolumn{3}{c|}{\textbf{Multi-Hop}} & \multicolumn{3}{c|}{\textbf{Open Domain}} & \multicolumn{3}{c|}{\textbf{Temporal}} & \multicolumn{3}{c}{\textbf{Overall}} \\
\cmidrule(lr){3-5} \cmidrule(lr){6-8} \cmidrule(lr){9-11} \cmidrule(lr){12-14} \cmidrule(lr){15-17}
& & F1 & BLEU & LLM Score & F1 & BLEU & LLM Score & F1 & BLEU & LLM Score & F1 & BLEU & LLM Score & F1 & BLEU & LLM Score \\
\midrule

\multirow{8}{*}{\rotatebox[origin=c]{90}{\textbf{gpt-4o-mini}}} 
& FullContext & 0.294 & 0.220 & 0.804 & 0.247 & 0.168 & 0.532 & 0.192 & 0.146 & 0.604 & 0.386 & 0.269 & 0.903 & 0.328 & 0.232 & 0.790 \\
\cmidrule(lr){2-17}
& LangMem & 0.388 & 0.331 & 0.614 & 0.335 & 0.239 & 0.524 & 0.294 & 0.235 & 0.476 & 0.319 & 0.262 & 0.249 & 0.358 & 0.294 & 0.513 \\
& Mem0 & 0.444 & 0.377 & 0.681 & 0.343 & 0.252 & 0.603 & 0.271 & 0.194 & 0.406 & 0.444 & 0.376 & 0.504 & 0.415 & 0.342 & 0.613 \\
& RAG & 0.222 & 0.186 & 0.320 & 0.186 & 0.117 & 0.313 & 0.190 & 0.135 & 0.326 & 0.195 & 0.157 & 0.237 & 0.208 & 0.164 & 0.302 \\
& Zep & 0.397 & 0.337 & 0.632 & 0.275 & 0.193 & 0.505 & 0.229 & 0.157 & 0.396 & 0.448 & 0.381 & 0.589 & 0.375 & 0.309 & 0.585 \\
& Nemori & 0.544 & 0.432 & \textbf{0.821} & 0.365 & 0.256 & 0.653 & 0.208 & 0.151 & 0.448 & 0.567 & 0.466 & 0.710 & 0.495 & 0.385 & 0.744 \\
& \textbf{Ours} & 0.297 & 0.223 & 0.765 & 0.410 & 0.296 & \textbf{0.813} & 0.201 & 0.164 & \textbf{0.645} & 0.410 & 0.296 & \textbf{0.888} & 0.377 & 0.274 & \textbf{0.835} \\
& \textbf{\textsubscript{\! \scriptsize \textnormal{+ CONSOLIDATION}}} & 0.304 & 0.271 & 0.797 & 0.428 & 0.316 & 0.764 & 0.182 & 0.143 & 0.625 & 0.391 & 0.281 & 0.875 & 0.370 & 0.230 & 0.823 \\

\midrule

\multirow{8}{*}{\rotatebox[origin=c]{90}{\textbf{gpt-4.1-mini}}}
& FullContext & 0.325 & 0.246 & 0.893 & 0.334 & 0.217 & 0.878 & 0.267 & 0.225 & 0.781 & 0.407 & 0.290 & 0.948 & 0.368 & 0.263 & 0.913 \\
\cmidrule(lr){2-17}
& LangMem & 0.510 & 0.436 & 0.845 & 0.415 & 0.325 & 0.710 & 0.328 & 0.264 & 0.590 & 0.485 & 0.409 & 0.508 & 0.476 & 0.400 & 0.734 \\
& Mem0 & 0.486 & 0.420 & 0.714 & 0.401 & 0.303 & 0.682 & 0.237 & 0.177 & 0.479 & 0.392 & 0.332 & 0.569 & 0.435 & 0.365 & 0.663 \\
& RAG & 0.258 & 0.220 & 0.359 & 0.201 & 0.128 & 0.317 & 0.179 & 0.139 & 0.288 & 0.223 & 0.191 & 0.274 & 0.235 & 0.192 & 0.329 \\
& Zep & 0.455 & 0.400 & 0.669 & 0.305 & 0.204 & 0.537 & 0.242 & 0.193 & 0.438 & 0.239 & 0.200 & 0.602 & 0.369 & 0.309 & 0.616 \\
& Nemori & 0.588 & 0.515 & 0.849 & 0.417 & 0.319 & 0.751 & 0.258 & 0.193 & 0.510 & 0.577 & 0.502 & 0.776 & 0.534 & 0.456 & 0.794 \\
& \textbf{Ours} & 0.298 & 0.223 & 0.858 & 0.312 & 0.199 & 0.872 & 0.249 & 0.205 & \textbf{0.697} & 0.400 & 0.282 & \textbf{0.952} & 0.354 & 0.249 & \textbf{0.902} \\
& \textbf{\textsubscript{\! \scriptsize \textnormal{+ CONSOLIDATION}}} & 0.298 & 0.223 & \textbf{0.861} & 0.309 & 0.208 & \textbf{0.878} & 0.248 & 0.203 & 0.666 & 0.391 & 0.275 & 0.934 & 0.348 & 0.246 & 0.892 \\
\bottomrule
\end{tabular}
}
\caption{Performance on the LoCoMo dataset using LLM-as-a-judge, F1 and BLEU scores. Baselines LangMem, Mem0, RAG, Zep and Nemori were taken from \cite{nan2025nemori}. Bold indicates the best among memory methods.}
\label{tab:main_locomo_full}
\vspace{-3mm}
\end{table*}

\subsection{Memory Consolidation}
As interactions accumulate, the memory store grows monotonically, increasing storage costs, retrieval overhead, and introduction of noise. To address this, we introduce a periodic clustering-based consolidation mechanism that groups semantically similar memories using HDBSCAN~\cite{campello2013density}, enabling the merging of evolving concepts, resolution of conflicting information, and de-duplication of redundant entries across sessions. As shown in Figure~\ref{fig:architecture}, all memories in a cluster are fed to an LLM which then decides whether to merge or delete memories. Formally,
let $\mathcal{M} = \{ m_1, m_2, \dots, m_N \}$ denote the set of stored memory items accumulated from user interactions, where each memory $m_i$ is represented by a semantic embedding $\mathbf{e}_i \in \mathbb{R}^d$ and associated metadata (e.g., timestamp, type).
We perform unsupervised clustering over $\mathcal{M}$, yielding a set of clusters
\begin{equation}
\mathcal{C} = \{ C_1, C_2, \dots, C_K \}, \quad C_k \subseteq \mathcal{M},
\end{equation}
where each cluster $C_k$ groups semantically similar memories corresponding to an evolving concept. For each cluster $C_k$, we apply a consolidation operator $\Phi(\cdot)$ that maps the cluster to a reduced memory set, $\Phi(C_k) \rightarrow \mathcal{M}_k^{\prime},$ via the following operations.
\begin{itemize}
    \item \textbf{MERGE}: merging memories in $C_k$ that are redundant or complementary into a single piece of consolidated memory, preserving salient information while reducing duplication:
    \begin{equation}
    \tilde{\mathcal{M}}_k = \operatorname{merge}(C_k)
    \end{equation}
    \item \textbf{DELETE}: removing a piece of memory $m_i \in C_k$ that is outdated, conflicting, or superseded:
    \begin{equation}
        \mathcal{M}_k^{\prime} = \tilde{\mathcal{M}}_k \setminus \{ m_i \}
    \end{equation}
\end{itemize}

After applying both operations, the consolidated memory store is defined as
\begin{equation}
    \mathcal{M}^{\prime} = \bigcup_{k=1}^{K} \mathcal{M}_k^{\prime}
\end{equation}
yielding a compact, denoised memory set.

\subsection{Memory Retrieval \& Question Answering}
\label{sec:retrieval}

At query time, we perform independent top-$k$ semantic similarity searches over the
\epimem{} and \structmem{} stores separately.
For \epimem{}, each episode's \textbf{details} field is indexed, yielding
one vector per episode.
For \structmem{}, each tuple is indexed as a natural-language string
concatenating its components (e.g., \textit{``Subject, Relation,
Object''}), yielding one vector per triplet.
The two stores are searched independently, returning the top-$k$ episodes from \epimem{} and the top-$2k$ tuples from \structmem{}. The episodic memories are more verbose than relational tuples, so within the same budget approximately twice as many relational entries can be retrieved.
We set $k=10$ since performance plateaus beyond this value (see the budget analysis in Section~\ref{sec:retrieval_budget}).

\begin{table*}[t] 
\centering
\small 
\setlength{\tabcolsep}{15pt}
\vspace{2mm}
\begin{tabular}{lccc}
\toprule
\textbf{Question Type} & \textbf{Full-context} & \textbf{Nemori} & \textbf{Ours} \\
& \scriptsize (101K tokens) & \scriptsize (3.7--4.8K tokens) & \scriptsize (3--6.9K tokens) \\
\midrule
single-session-preference & 0.167 & \textbf{0.867} & 0.500 \\
single-session-assistant  & \textbf{0.982} & 0.929 & \textbf{0.982} \\
temporal-reasoning        & 0.602 & 0.722 & \textbf{0.858} \\
multi-session             & 0.511 & 0.556 & \textbf{0.801} \\
knowledge-update          & 0.769 & 0.795 & \textbf{0.902} \\
single-session-user       & 0.857 & 0.900 & \textbf{0.953} \\
\midrule
Average          & 0.656 & 0.746 & \textbf{0.855} \\
\bottomrule
\end{tabular}
\caption{Detailed performance comparison on \longmemeval{} (gpt-4.1-mini) across question types.}
\label{tab:longmem_main}
\vspace{-2mm}
\end{table*}

\begin{table}[ht]
\centering
\small
\begin{tabular}{lrr}
\toprule
\textbf{Method} & \textbf{LLM Score} & \textbf{Tokens} \\
\midrule
FullContext & 0.790 & 21,788 \\
RAG-4096    & 0.302 & 3,430  \\
LangMem     & 0.513 & 125    \\
Zep         & 0.585 & 2,247  \\
Mem0        & 0.613 & 1,027  \\
Nemori      & 0.744 & 2,745 \\
\textbf{Ours}        & \textbf{0.835} & 2,738 \\
\bottomrule
\end{tabular}
\caption{Performance and efficiency comparison on LoCoMo dataset with gpt-4o-mini.}
\label{tab:efficiency_comparison}
\vspace{-5mm}
\end{table}

\section{Experiments}
\label{sec:setup}
We evaluate our framework on two long-horizon dialogue benchmarks:
\textsc{LoCoMo}~\cite{maharana2024evaluating}, featuring 1,540 QA pairs from 10 distinct dialogues across four reasoning categories: \textit{Single-Hop}, \textit{Multi-Hop}, \textit{Temporal}, and \textit{Open-domain}. On average, the conversations in LoCoMo span multiple sessions with 24K tokens; and \textsc{LongMemEval-S}~\cite{wu2024longmemeval}, a substantially more challenging evaluation setting, comprising 500 meticulously curated questions embedded within synthetic but realistic chat histories that average 115K tokens per question.

\subsection{Experiment Setup}

\paragraph{Baselines.}
We compare with state-of-the-art memory frameworks, including \textsc{FullContext} (all sessions placed in context) \textsc{RAG-4096} (sessions chunked into 4096 token segments), \textsc{LangMem}~\cite{chase2022langchain}, \textsc{Mem0}~\cite{chhikara2025mem0}, \textsc{Zep}~\cite{rasmussen2025zep}, and \textsc{Nemori}~\cite{nan2025nemori}.

\paragraph{Implementation.}
We report results where memory curation, consolidation and answer generation are backed by \texttt{gpt-4.1-mini} and \texttt{gpt-4o-mini} models. We use \texttt{text-embedding-3-small} for vector embedding and retrieval. For clustering we have used UMAP~\cite{mcinnes2018umap} for dimension reduction and density based HDBSCAN~\cite{campello2013density} for clustering. Detailed hyperparameters are in Appendix~\ref{app:implementation_details}.

\subsection{Results}
\label{sec:results}

\paragraph{Main Results.}

Table~\ref{tab:main_locomo_full} shows the performance on the \locomo{} benchmark. Our system consistently outperforms all baselines when using \texttt{gpt-4o-mini}, achieving a score of $83.5\%$. Notably, this surpasses the \texttt{FullContext} baseline, empirically validating that the addition of long context can lead to performance degradation. With \texttt{gpt-4.1-mini}, our approach reaches $90.2\%$, a significant $6.7\%$ absolute improvement over the \texttt{gpt-4o-mini} results. While this score is slightly below the \texttt{FullContext} performance of $91.3\%$, our system remains highly competitive while utilizing $87\%$ fewer tokens.

The most significant gains are observed in \textbf{Multi-Hop} and \textbf{Temporal} reasoning tasks, which directly reflect the strengths of our dual-memory setup. Episodic memory preserves interactions as a timestamped timeline, allowing the model to track the order of events for temporal questions. Meanwhile, structured relational memories capture complex entity-relation dependencies, facilitating effective multi-hop inference over distinct facts.
Appendix~\ref{app:component_analysis} provides an ablation study on the contribution of each component. The results demonstrate that the most substantial improvements (+6.3 points in Multi-Hop and +6.8 points in Temporal reasoning on \texttt{gpt-4o-mini}) emerge due to interaction between the two memory types.

Table~\ref{tab:longmem_main} presents the results for the \longmemeval{} benchmark. Our method attains an overall score of $85.5\%$, consistently surpassing both the \textsc{FullContext} baseline and \textsc{Nemori} across nearly all evaluation categories.

\paragraph{Efficiency.}

In Table~\ref{tab:efficiency_comparison}, we compare our system against the baselines in terms of the average number of tokens consumed per question. Our approach uses comparable tokens to \textsc{Nemori}, while achieving a $9.1$ point improvement in accuracy. 
Furthermore, as in Table~\ref{tab:consolidation_sensitivity}, our consolidation method reduces the size of the memory store by $16.34\%$ with only 1\% degradation in performance, demonstrating effective memory compression without much loss in accuracy.

\section{Analysis and Ablations}
\label{sec:ablation}

We analyse the sensitivity of the system to its two main hyperparameters and
its robustness across backbone LLMs. The contribution of each individual
memory type is reported in Appendix~\ref{app:component_analysis}. All
experiments in this section use \locomo{}.

\subsection{Generalization Across Models}
\label{sec:backbone_eval}

To assess robustness across backbones, we additionally evaluate a stronger
proprietary model (\texttt{gpt-5.4}) and a smaller open-weight model
(\texttt{Qwen3-8B}), shown in Table~\ref{tab:backbone_eval}.

\begin{table}[ht]
\centering
\resizebox{0.98\columnwidth}{!}{
\begin{tabular}{lccccc}
\toprule
\textbf{Model} & \textbf{Overall} & \textbf{S-Hop} & \textbf{M-Hop} & \textbf{Open} & \textbf{Temp.} \\
\midrule
gpt-5.4       & 0.906 & 0.865 & 0.903 & 0.729 & 0.941 \\
gpt-4.1-mini  & 0.902 & 0.858 & 0.872 & 0.697 & 0.952 \\
gpt-4o-mini   & 0.835 & 0.765 & 0.813 & 0.645 & 0.888 \\
Qwen3-8B      & 0.762 & 0.748 & 0.629 & 0.552 & 0.841 \\
\bottomrule
\end{tabular}
}
\caption{Performance of our framework with different backbone LLMs on \locomo{}.}
\label{tab:backbone_eval}
\vspace{-3mm}
\end{table}

The proposed memory system generalizes across backbones of varying capacity.
Notably, \texttt{gpt-4.1-mini} performs comparably to \texttt{gpt-5.4},
supporting our claim that the primary challenge lies in designing an
effective memory system rather than relying on increased model capacity. The
open-weight \texttt{Qwen3-8B} backbone, despite being substantially smaller,
still attains 76.2\% overall accuracy, confirming that our method is not
restricted to stronger models.

\subsection{Effect of Retrieval Budget}
\label{sec:retrieval_budget}

\begin{figure}[t]
\centering
\includegraphics[width=0.9\columnwidth]{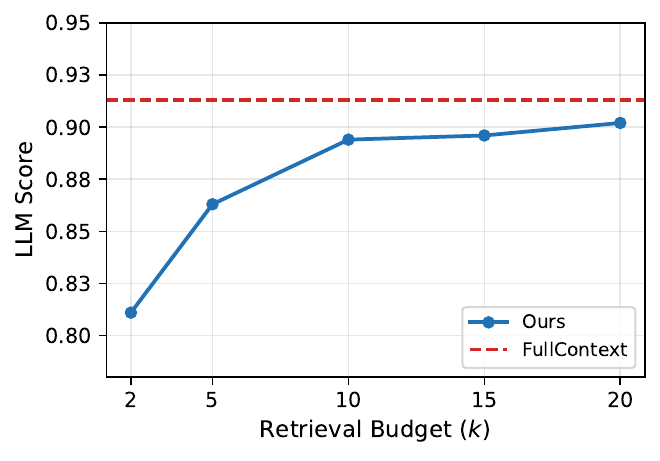}
\vspace{-6mm}
\caption{Effect of retrieval budget $k$ on LLM Score with \texttt{gpt-4.1-mini}. Performance plateaus at $k{=}10$.}
\label{fig:retrieval_budget}
\vspace{-3mm}
\end{figure}

To study how much retrieved evidence the system needs, we vary the number of
retrieved memories $k$ (both episodic and relational) and report overall LLM
Score with \texttt{gpt-4.1-mini} in Figure~\ref{fig:retrieval_budget}.

Performance increases sharply from $k{=}2$ to $k{=}10$ and then plateaus,
with diminishing returns of $<$1 point between $k{=}10$ and $k{=}20$.
Notably, even at $k{=}2$ the system retains 0.811 score, indicating that
retrieval is reasonably precise rather than relying on a large budget.

\subsection{Effect of Consolidation Cluster Size}
\label{sec:consolidation_sensitivity}

We conduct a sensitivity analysis on the primary clustering hyperparameter,
\texttt{min\_cluster\_size} (mcs), which controls the minimum number of
memory points required to form a cluster. We evaluate
mcs$\in\{2, 5, 10, 20\}$ with \texttt{gpt-4.1-mini} in
Table~\ref{tab:consolidation_sensitivity}.

\begin{table}[ht]
\centering
\small
\begin{tabular}{rccc}
\toprule
\textbf{mcs} & \textbf{Compression} & \textbf{LLM Score} & \textbf{Clusters} \\
\midrule
2   & 30.9\% & 0.876 & 2208 \\
5   & 21.7\% & 0.896 & 693 \\
10  & 16.3\% & 0.892 & 325 \\
20  & 9.6\%  & 0.898 & 144 \\
\bottomrule
\end{tabular}
\caption{Effect of \texttt{min\_cluster\_size} on consolidation.}
\label{tab:consolidation_sensitivity}
\vspace{-3mm}
\end{table}

Performance is largely stable for mcs$\in[5,20]$, with LLM Score ranging from
0.892--0.898, all within 1 point of the no-consolidation baseline (0.902). At
mcs$=2$, performance drops by 2.6 points, which we attribute to
over-aggressive consolidation where small groups of similar embeddings are
prematurely merged. As mcs increases, memory compression decreases
monotonically from 30.9\% to 9.6\%, while the number of clusters (and
corresponding LLM consolidation calls) is reduced by up to 15$\times$. This
highlights a favorable trade-off: moderate cluster sizes (mcs$\in[5,10]$)
achieve meaningful compression (16--22\%) with negligible accuracy loss.

\section{Conclusion}
\label{sec:conclusion}

We presented a \sysname{} system for agents that
curates and retrieves both \epimem{} and \structmem{}{}, and conducted an
empirical study across \locomo{} and \longmemeval{}.
Our results demonstrate strong performance combining both memory types, and improved efficiency with memory consolidation without much loss in performance.

\section{Limitations}
While our proposed memory system demonstrates strong performance, several areas remain for further refinement.
First, our current evaluation focuses exclusively on textual conversational data.
Although the system is modality agnostic, extending it to support non-textual modalities such as vision and audio remains an important direction for future work.
Second, our experiments reveal that memory curation quality can vary substantially across different backbone LLMs, as observed shifting from \texttt{gpt-4.1-mini} to \texttt{gpt-4o-mini} resulted in a 6.7 percentage point drop.
This sensitivity reflects a broader challenge faced by existing agentic memory systems, where memory curation and consolidation are inherently influenced by model-specific reasoning and representation biases.
Developing memory mechanisms that are robust and consistent across diverse model families remains an open and important research problem.

\bibliography{custom}

\appendix
\onecolumn
\label{sec:appendix}

\section{Prompt used for Episodic Memory Curation}
\label{app:episodic_memory_prompt}
\begin{figure*}[!ht]
\centering
\begin{tcolorbox}[
    fonttitle=\small\bfseries\sffamily,
    colback=gray!2,
    colframe=gray!60,
    sharp corners,
    boxrule=0.6pt,
    left=15pt, right=15pt, top=10pt, bottom=10pt
]
    \small
    \noindent\textbf{Role:} You are an Episodic Memory Curator. Your job is to extract HOLISTIC episodes from conversations capturing complete narratives rather than fragmented facts.

    \vspace{0.5em}
    \noindent\textbf{TASK:} Extract CONSOLIDATED episodic memories from the conversation below. Focus on creating coherent, comprehensive episodes that capture the full context of events.

    \vspace{0.8em}
    \noindent\texttt{Current Date: \{session\_date\}} \\
    \noindent\texttt{CONVERSATION: \{session\_text\}}

    \vspace{0.8em}
    \hrule
    \vspace{0.8em}

    \noindent\textbf{EXTRACTION RULES:}

    \begin{enumerate}
        \item \textbf{WHAT TO CAPTURE (be exhaustive):}
        \begin{itemize}
            \item Events: meetings, trips, celebrations, appointments, activities.
            \item Plans and scheduled activities (with exact dates if mentioned).
            \item Preferences, opinions, and subjective feelings (e.g., ``X loves hiking'').
            \item Biographical facts: family details, careers, hobbies, pets.
            \item Specific quantities: counts, frequencies, durations.
            \item Advice given, specific phrases, exact quotes worth remembering.
            \item Relationships between people mentioned.
            \item Past events mentioned (e.g., ``last year we saw X'').
            \item \textit{NOT: Casual greetings or meta-conversation.}
        \end{itemize}

        \vspace{0.5em}
        \item \textbf{TEMPORAL PRECISION (CRITICAL):}
        \begin{itemize}
            \item The ``Current Date'' is WHEN THIS CONVERSATION HAPPENED.
            \item For relative expressions, show your arithmetic explicitly:
            \begin{itemize}
                \item ``last Saturday'' $\rightarrow$ count back to most recent Saturday before Current Date.
                \item ``next month'' $\rightarrow$ the calendar month after Current Date's month.
                \item ``yesterday'' $\rightarrow$ Current Date minus 1 day.
            \end{itemize}
            \item Use YYYY-MM-DD, YYYY-MM, or YYYY format. If no info, use null.
        \end{itemize}

        \vspace{0.5em}
        \item \textbf{DETAIL PRESERVATION (CRITICAL):}
        \begin{itemize}
            \item Capture EXACT names, numbers, places, titles, and species.
            \item Preserve exact sentiments and verbatim frequencies.
            \item Prefer granular episodes over fewer big ones.
        \end{itemize}

        \vspace{0.5em}
        \item \textbf{MEDIA INFORMATION}
        \begin{itemize}
            \item Preserve full caption text; record visible text (signs, posters).
            \item Describe specific objects, scenes, or people visible in the image.
        \end{itemize}

        \vspace{0.5em}
        \item \textbf{OUTPUT FORMAT:} 
        Return a JSON array of objects with this exact structure:
        \begin{quote}
        \begin{verbatim}
  {
    "title": "Short descriptive title (3-7 words)",
    "summary": "One-sentence description",
    "details": "Concise specifics. Names, numbers, quotes.",
    "timestamp": "ISO date string (YYYY-MM-DD, YYYY-MM, YYYY) or null"
  }
        \end{verbatim}
        \end{quote}
    \end{enumerate}

    \vspace{0.8em}
    \hrule
    \vspace{0.8em}
    \noindent\textbf{OUTPUT JSON ONLY (No markdown, no explanations)}
\end{tcolorbox}
\label{app:prompt_episodic_curation}
\end{figure*}

\newpage
\section{Prompt used for Relational Memory Curation}
\label{app:relational_memory_prompt}
\begin{figure*}[!ht]
\centering
\begin{tcolorbox}[
    fonttitle=\small\bfseries\sffamily,
    colback=white,
    colframe=gray!60,
    sharp corners,
    boxrule=0.6pt,
    left=12pt, right=12pt, top=10pt, bottom=10pt
]
    \footnotesize
    \noindent\textbf{Role:} You are an expert Relational Memories Curator. Extract ALL meaningful facts from this conversation as triplets.

    \vspace{0.5em}
    \noindent\textbf{TASK:} Extract relational memories from the conversation below.

    \vspace{0.5em}
    \noindent\texttt{Current Date: \{session\_date\}} \\
    \noindent\texttt{CONVERSATION: \{session\_text\}}
    \vspace{0.8em}
    \hrule
    \vspace{0.8em}

    \noindent\textbf{EXTRACTION RULES:}

    \begin{enumerate}
        \item \textbf{WHAT TO EXTRACT (be exhaustive)}
        \begin{itemize}
            \item Events: meetings, trips, celebrations, appointments, activities.
            \item Plans and scheduled activities (with exact dates if mentioned).
            \item Preferences, opinions, and subjective feelings (e.g., ``X loves hiking'').
            \item Biographical facts: family details, careers, hobbies, pets.
            \item Specific quantities: counts, frequencies, durations.
            \item Advice given, specific phrases, exact quotes worth remembering.
            \item Relationships between people mentioned.
            \item Past events mentioned (e.g., ``last year we saw X'').
            \item \textit{NOT: Casual greetings or meta-conversation.}
        \end{itemize}

        \vspace{0.5em}
        \item \textbf{ENTITY RULES:}
        \begin{itemize}
            \item \textbf{RESOLVE PRONOUNS:} Map ``I'', ``me'', ``my'' to the speaker's actual name. Map ``you'', ``your'' to the other person's name.
            \item \textbf{VERIFY ACCURACY:} If the text says ``my dog Luna and my cat Oliver'', then Luna=dog, Oliver=cat. Do NOT swap species/names.
        \end{itemize}

        \vspace{0.5em}
        \item \textbf{TEMPORAL PRECISION (CRITICAL --- show your date math):}
        \begin{itemize}
            \item Current Date is WHEN THIS CONVERSATION HAPPENED.
            \item For ALL relative time references, resolve to absolute dates with explicit arithmetic:
            \begin{itemize}
                \item ``last Saturday'' $\rightarrow$ find the Saturday before Current Date $\rightarrow$ specific YYYY-MM-DD
                \item ``next month'' $\rightarrow$ month after Current Date's month $\rightarrow$ YYYY-MM
                \item ``last year'' $\rightarrow$ Current Date year - 1 $\rightarrow$ YYYY
                \item ``yesterday'' $\rightarrow$ Current Date - 1 day $\rightarrow$ YYYY-MM-DD
                \item ``10 years ago'' $\rightarrow$ Current Date year - 10 $\rightarrow$ YYYY
                \item ``two weekends ago'' $\rightarrow$ Current Date - 14 days $\rightarrow$ find that weekend $\rightarrow$ YYYY-MM-DD
            \end{itemize}
            \item \textbf{ALWAYS try to infer a date} --- even an approximate month/year is better than null.
        \end{itemize}

        \vspace{0.5em}
        \item \textbf{PERSISTENCE FILTERING:}
        \begin{itemize}
            \item \textbf{KEEP:} Everything from categories 1a-1j above.
            \item \textbf{DISCARD ONLY:} Pure greetings (``Hey!'', ``Hi!''), meta-conversation (``tell me more''), and filler.
            \item \textbf{When in doubt, KEEP IT} --- it's better to have an extra triplet than to miss one.
        \end{itemize}

        \vspace{0.5em}
        \item \textbf{OUTPUT FORMAT (JSON List):} 
        Return a JSON list where each object follows this exact structure:
        \begin{quote}
        \begin{verbatim}
{
  "subject": "Canonical Entity Name",
  "relation": "predicate_snake_case",
  "object": "Specific value (names, numbers, adjectives)",
  "when": "YYYY-MM-DD or YYYY-MM or YYYY or null",
  "context": "source text for grounding",
}
        \end{verbatim}
        \end{quote}
    \end{enumerate}

    \vspace{0.8em}
    \hrule
    \vspace{0.8em}
    \noindent\textbf{OUTPUT JSON ONLY (No markdown, no explanation)}
\end{tcolorbox}
\label{app:prompt_relation_curation}
\end{figure*}

\twocolumn
\section{Implementation details}
\label{app:implementation_details}
The consolidation pipeline follows a two-phase architecture inspired by BERTopic~\cite{grootendorst2022bertopic}. In the first phase, all stored memories with their pre-computed 1536-dimensional embeddings produced by OpenAI's text-embedding-3-small model are projected into a lower-dimensional space using UMAP~\cite{mcinnes2018umap} with cosine distance, 10 neighbors, 5 target components, and minimum distance of zero.  The reduced representations are then clustered using HDBSCAN~\cite{campello2013density} with Euclidean distance, Excess of Mass cluster selection, minimum cluster size of 10, and minimum samples of 1. 

\section{Memory Component Analysis}
\label{app:component_analysis}

To understand the contributions of each component, we evaluate the
episodic-only, relational-only, full hybrid, and consolidated variants on
\locomo{}. Results are shown in Table~\ref{tab:ablation_components}.

\begin{table}[ht]
\centering
\resizebox{0.98\columnwidth}{!}{
\begin{tabular}{lccccc}
\toprule
\textbf{Component} & \textbf{Overall} & \textbf{S-Hop} & \textbf{M-Hop} & \textbf{Open} & \textbf{Temp.} \\
\midrule
\multicolumn{6}{l}{\textit{gpt-4o-mini}} \\
Episodic only    & 0.776 & 0.765 & 0.710 & 0.645 & 0.820 \\
Relational only  & 0.730 & 0.680 & 0.750 & 0.572 & 0.757 \\
Ours             & 0.835 & 0.765 & 0.813 & 0.645 & 0.888 \\
+ Consolidation  & 0.823 & 0.797 & 0.764 & 0.625 & 0.875 \\
\midrule
\multicolumn{6}{l}{\textit{gpt-4.1-mini}} \\
Episodic only    & 0.871 & 0.822 & 0.847 & 0.687 & 0.918 \\
Relational only  & 0.837 & 0.781 & 0.831 & 0.625 & 0.882 \\
Ours             & 0.902 & 0.858 & 0.872 & 0.697 & 0.952 \\
+ Consolidation  & 0.892 & 0.861 & 0.878 & 0.666 & 0.934 \\
\bottomrule
\end{tabular}
}
\caption{Memory component ablation on \locomo{} with \texttt{gpt-4o-mini} and \texttt{gpt-4.1-mini}.}
\label{tab:ablation_components}
\end{table}

The combined system consistently outperforms either single memory type.
Substantial gains over the stronger individual component appear on
\textbf{Multi-Hop} (+6.3 points with \texttt{gpt-4o-mini}, +2.5 points with
\texttt{gpt-4.1-mini}) and \textbf{Temporal} (+6.8 points with
\texttt{gpt-4o-mini}, +3.4 points with \texttt{gpt-4.1-mini}), where the two
memory types are complementary. On Single-Hop and Open-Domain, the hybrid
matches the best single component with \texttt{gpt-4o-mini} and further
improves performance by 3.6 and 1.0 points, respectively, with
\texttt{gpt-4.1-mini}.

\section{Extended Comparison with Prior Work}
\label{app:extended_related}

Prior memory systems differ from ours along three axes: \emph{representation}, \emph{curation strategy}, and \emph{operational cost}. We summarize the key contrasts below.

\paragraph{Representation.}
Zep~\cite{rasmussen2025zep} maintains a temporal knowledge graph with entity nodes, fact edges, and validity intervals, enabling graph traversal at the cost of entity resolution, edge deduplication, and conflict detection. A-Mem~\cite{amem} organizes memories as Zettelkasten-style notes with evolving links between entries, while H-Mem~\cite{hmem} relies on hierarchical indexing for routing. In contrast, we use two complementary \emph{flat} stores: episodic records that preserve narrative context and relational tuples retrieved by semantic similarity. This sacrifices native graph traversal but avoids the overhead of graph construction and hierarchy maintenance.

\paragraph{Curation strategy.}
Mem0~\cite{chhikara2025mem0} processes each turn individually with add/update/delete/no-op operations, which fragments inter-turn context. Nemori~\cite{nan2025nemori} adaptively distills semantic knowledge based on prediction error against an anticipatory schema, focusing on \emph{what} to store. Our method instead targets \emph{how} memory should be represented: we perform a single session-level curation pass that produces episodic and relational entries, and we apply offline clustering-based consolidation to merge redundant memories and remove outdated ones.

\paragraph{Operational cost.}
Turnwise systems such as Mem0 and A-Mem scale LLM calls linearly with conversation length, and dynamic retrieval schemes incur additional LLM rounds at query time. Our session-level curation amortizes extraction across many turns, and a fixed top-$k$ retrieval avoids extra LLM calls. Consolidation runs offline and can be triggered infrequently (e.g., every 10-20 sessions), keeping long-horizon operation tractable while preserving QA accuracy.

\end{document}